\pdfoutput=1
\documentclass[11pt]{article}

\usepackage[]{acl}

\usepackage{times}
\usepackage{latexsym}

\usepackage[T1]{fontenc}
\usepackage[utf8]{inputenc}

\usepackage{microtype}

\usepackage{inconsolata}

\usepackage{graphicx}

\usepackage{booktabs}
\usepackage{multirow}
\usepackage{subfigure}
\usepackage{amsmath}
\usepackage{algorithm} 
\usepackage{algorithmic}
\usepackage{bbding}

\title{Leveraging Grammar Induction for Language Understanding and Generation}

\author{Jushi Kai \quad Shengyuan Hou \quad Yusheng Huang \quad Zhouhan Lin\Thanks{~Corresponding author} \\
\textsuperscript{} Shanghai Jiao Tong University \\
\textsuperscript{} json.kai@sjtu.edu.cn \quad lin.zhouhan@gmail.com \\}

\begin{document}
\maketitle
\begin{abstract}
Grammar induction has made significant progress in recent years. However, it is not clear how the application of induced grammar could enhance practical performance in downstream tasks.
In this work, we introduce an unsupervised grammar induction method for language understanding and generation. We construct a grammar parser to induce constituency structures and dependency relations, which is simultaneously trained on downstream tasks without additional syntax annotations. The induced grammar features are subsequently incorporated into Transformer as a syntactic mask to guide self-attention. We evaluate and apply our method to multiple machine translation tasks and natural language understanding tasks. Our method demonstrates superior performance compared to the original Transformer and other models enhanced with external parsers. Experimental results indicate that our method is effective in both from-scratch and pre-trained scenarios. Additionally, our research highlights the contribution of explicitly modeling the grammatical structure of texts to neural network models.\footnote{We release our code at \url{https://github.com/LUMIA-Group/Leveraging-Grammar-Induction}.} %\footnote{We will release our code for reproducibility later.}
\end{abstract}

\section{Introduction}

% The models for language comprehension fall into two main categories. One is serialized neural network models, especially those under seq2seq (\citealp{fairseq}) architecture, like LSTM (\citealp{lstm}) and Transformer (\citealp{transformer}). They use serialized multi-layer neural network structure, which can implicitly learn the syntactic and semantic information of the language, rather than explicitly modeling the grammatical structure. The other one is traditional syntax parsers (\citealp{parser}; \citealp{pearl}), which essentially rely on supervised learning to explicitly model the grammatical structure of textual data.

Neural network models, like Transformer (\citealp{transformer}), RoBERTa (\citealp{roberta}), and GPTs (\citealp{gpt}), have gained widespread adoption in various natural language processing tasks. These models can generate desired answers on different tasks and show strong language understanding ability on multiple datasets. However, they give up explicit parsing of the specific syntactic structure of the text data and cannot effectively establish structured and interpretable language understanding models (\citealp{vqa-regularization}; \citealp{vqa-prior}; \citealp{structural-probe}; \citealp{absa}). This limitation has emerged as one of the bottlenecks for neural network models to understand natural language deeply.

To this end, researchers attempt to take advantage of traditional syntax parsers \cite{parser, pearl} to identify grammatical components within textual data and utilize them in subsequent processing steps. These parsers are built upon established linguistic frameworks and regulations. There has been growing interest in investigating the impact of syntax on neural network models and improving them through the lens of grammar induction \cite{st-nmt, syntax-bert, sla-bert, distance-tranformer}.

While these techniques have exhibited enhanced efficacy, they still grapple with two primary challenges. Firstly, these approaches \cite{st-nmt, syntax-bert, distance-tranformer} usually depend on external parsers to obtain additional specialized annotations, which are expensive and not time-efficient. Their performance is greatly influenced by the choice of external parsers and may not be universally applicable across all languages. The other problem is that these approaches only focus on from-scratch scenarios \cite{nsd-nmt, st-nmt} or pre-trained scenarios \cite{sla-bert, tag-bert, absa-bert}). They cannot share consistent improvements. This discrepancy between the two model types engenders a predicament where these methodologies are tailored exclusively to one of the scenarios.

\begin{figure*}[ht]
    \centering
    \includegraphics[width=1\linewidth]
    {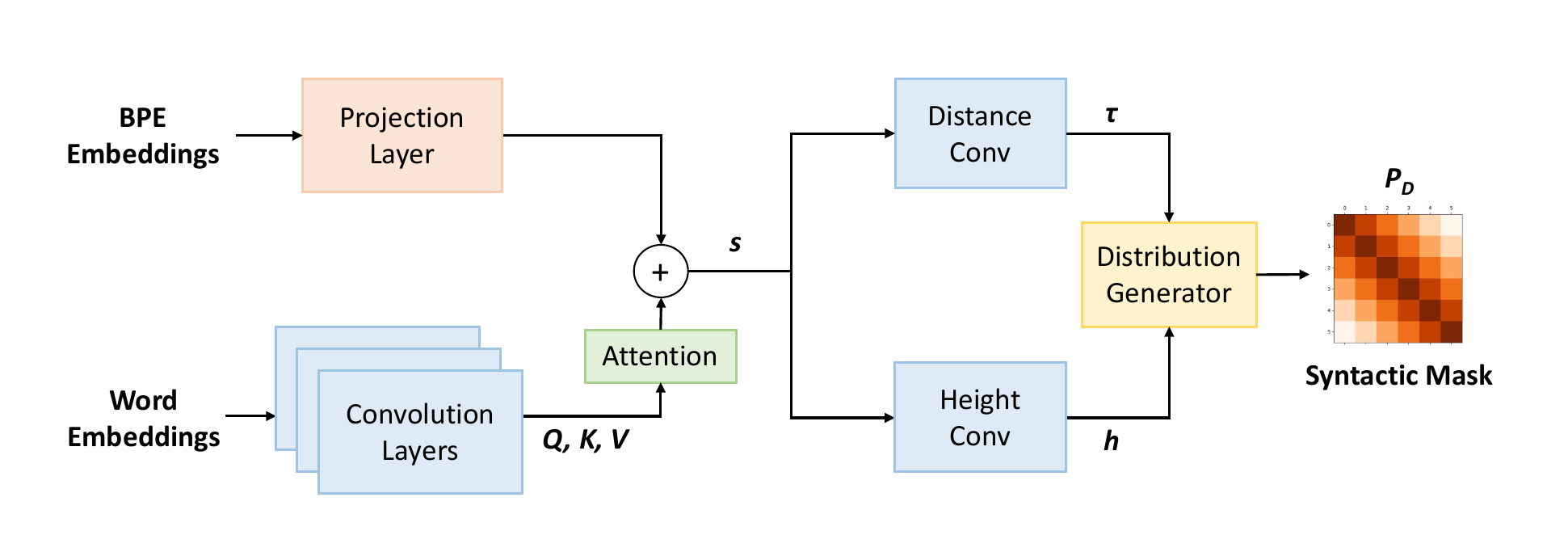}
    \caption{The pipeline for the construction of our syntactic mask. Word embeddings and BPE embeddings are utilized to induce the intermediate grammar features $s$, which are subsequently used to derive the syntactic distance $\tau$ and height $h$. The two vectors are leveraged to estimate the dependency distribution for the sentence, and generate the syntactic mask $P_D$. The mask is then employed to guide the self-attention mechanism within the encoders. These parsing modules are integrated into the Transformer model and trained together in downstream tasks.}
    \label{parser-fig}
\end{figure*}

In this paper, we introduce a novel method to induce grammar information for language understanding and generation, obviating the necessity for supplementary syntactic annotations. \textit{The self-induced grammar features are integrated into the Transformer and simultaneously learned during the training of downstream tasks}. Additionally, we devise BPE (Byte Pair Encoding) embeddings and trade-off loss functions to facilitate grammar induction. Experiments demonstrate the compatibility of our method in both the from-scratch and pre-trained scenarios. By strengthening the induction ability of the deep neural network model to the grammatical information, we can improve the model's understanding of natural language. Our method surpasses other external-parser-enhanced methods in machine translation and language understanding tasks, showcasing persistent efficacy and applicability.

\section{Preliminary}

We use syntactic distance \cite{PRPN} and height \cite{syntactic_height} to generate the dependency distribution among tokens. They are two feature vectors used to describe the constituency structure and dependency relations of a sequence of words.

\subsection{Syntactic Distance and Height}

\paragraph{Syntactic distance}
Syntactic distance is first proposed by \citet{PRPN} to model the syntactical proximity of adjacent constituents in a sentence. For a sentence $(w_1, \dots, w_n)$, syntactic distance $\tau_i$ quantifies the height of the lowest common ancestor for two consecutive words $w_i$ and $w_{i+1}$. The relative order of syntactic distances represents syntactic affinities between words. A smaller distance $\tau_i$ signifies that it is easier to communicate between $w_{1 \dots i}$ and $w_{i+1 \dots n}$.

\paragraph{Syntactic height}
Syntactic height is introduced by \citet{syntactic_height} to depict the syntactic status of words. In the dependency graph of a sentence, the syntactic height $h_i$ aims to capture the distance of the word $w_i$ to the root. The word with a higher syntactic height is more likely to stand closer to the root of the graph, containing more representative global information for the sentence.

\subsection{Dependency Distribution}
\label{sec:dist}

StructFormer \cite{structformer} proposes a two-stage method to calculate the estimation of the dependency distribution among tokens. They use syntactic distance and height to identify the smallest legal constituent for each token and the parent of the constituent. The procedure goes through the sentence to estimate the syntactic dependency of each token on the others. We leverage their estimation method to generate the \textbf{dependency distribution} for a sentence. The estimation procedure can be referred to in Appendix~\ref{sec:structformer}.

We introduce unsupervised grammar learning as inductive bias into Transformer for NLP downstream tasks. Our method achieves consistent improvement in both from-scratch and pre-trained scenarios.

\section{Method}

Alongside the data pipeline in the original Transformer, we construct a parser to derive grammar features and estimate a syntactic mask to guide the attention mechanism. The "parsing"\footnote{Despite no ground-truth parses involved in our method, the pipeline can be regarded as parsing to estimate dependency relations among tokens.} pipeline is delineated in Figure \ref{parser-fig}.

Grammar features $s$ are induced with convolution layers and a self-attention module working on word embeddings. Specifically, convolution layers can capture localized details within segments, while the self-attention module can furnish global information of the entire sequence. BPE embeddings are introduced so that grammar status can be shared among congenetic subwords. Utilizing grammar features $s$, syntactic distance $\tau$, and height $h$ can be derived through convolution layers. They are leveraged by the distribution generator to construct a syntactic mask $P_D$ that provides the dependency distribution among tokens in the encoder layers of Transformer.

\subsection{Syntactic Mask}

Following previous works (\citealp{PRPN}; \citealp{syntactic_height}; \citealp{structformer}), we quantify syntactic distance and height on the same scale. We use convolution layers and a self-attention module to induce grammar features $s_i$:
\begin{equation}
\begin{split}
    s_i &= \text{Attn}(\text{Conv}(x_{i-l}, \dots, x_{i+l})) \\
        & + \text{Proj}(x'_{i-l}, \dots, x'_{i+l})
\end{split}
\end{equation}
where $x_i$ is the embedding of the $i$-th token, and $2l+1$ is the kernel size of the convolution module. $x'_i$ is the BPE embedding, which will be introduced in Section \ref{sec:bpe}. $\text{Proj}(\cdot)$ is the projection layer composed of a Linear layer and a LayerNorm layer.

Then we use linear matrices ${W}_1^\tau, {W}_1^h$ and convolution modules ${W}_2^\tau$, ${W}_2^h$ to derive syntactic distance $\tau_i$ and height $h_i$:
\begin{equation}
    \tau_i=\mathbf{W}_1^\tau \tanh      \left(\mathbf{W}_2^\tau\left[\begin{array}{c}
        s_{i} \\
        s_{i+1}
\end{array}\right]\right)+b_1^\tau
\end{equation}
\begin{equation}
    h_i=\mathbf{W}_1^h \tanh \left(\mathbf{W}_2^h s_{i}+b_2^h\right)+b_1^h
\end{equation}

Syntactic distance measures the syntactic proximity of adjacent words, while syntactic height measures the syntactic status of the word itself. Therefore, the kernel size of ${W}_2^\tau$ and ${W}_2^h$ is 2 and 1 respectively.

As introduced in Section \ref{sec:dist}, syntactic distance and height are put into the distribution generator to calculate the estimation of the dependency distribution. It serves as a mask to adjust the weights in the self-attention:
\begin{equation}
\label{eq:estimation}
    P_D = \mathcal{F_P}(\tau, h)
\end{equation}
where the distribution generator is formalized as the estimation function $\mathcal{F_P}(\cdot)$. $P_D$ is the derived syntactic mask containing the dependency distributions.

\subsection{Syntax-giuded Attention}

The syntactic mask $P_D$ provides the probability of information transfer among tokens. It is used to guide self-attention heads in the encoder layers of Transformer.
\begin{equation}
\operatorname{Attention}(P_D, Q, K, V)=P_D \cdot \text{S}\left(\frac{Q K^T}{\sqrt{d}}\right) \cdot V
\end{equation}
where Q, K, V are query, key and value matrices and $d$ is the hidden dimension. $\text{S}(\cdot)$ is an activation function.

It should be noted that two different activation functions are employed in the scenarios of building from scratch and utilizing pre-trained models. In the from-scratch scenario, $\text{S}(\cdot)$ is a sigmoid function instead of the original softmax function so that $\text{S}(\frac{Q K^T}{\sqrt{d_k}})$ can indicate an independent probability of each token's attendance on each other.

In the pre-trained setting, we adhere to the use of the softmax function. Given our utilization of the official pre-trained models RoBERTa \cite{roberta} and the integration of our parser during fine-tuning, maintaining consistency in the activation function with the pre-training phase is more suitable and practical. In addition, the distribution weight is set to be $P_D+1$, where the syntactic mask plays an auxiliary role in reweighting attention. This is because RoBERTa has been pre-trained devoid of grammar induction. Directly imposing new constraints and altering the model training paradigm may not align seamlessly with its existing framework.

\subsection{BPE Embedding}
\label{sec:bpe}

When parsing, we compute dependency distributions for the tokens within sequences. Typically, the model receives subwords rather than complete words as input, owing to the data being preprocessed using BPE subword tokenization. Nonetheless, all subwords stemming from the same word should share an equivalent grammatical status within their respective sentences.

To make our parser aware of the grammar sharing among the congenetic subwords, we design BPE embeddings to represent the condition of word segmentation in the sequence. Tokens that remain intact are labeled as \textbf{0}, while subwords resulting from word segmentation are assigned \textbf{2} (with \textbf{1} reserved for padding). 
% An illustration of constructing a sentence's BPE embedding is depicted in Figure \ref{bpe-fig}. 
For instance, in the sentence "How could pay arrangements be redesigned to address these problems?", the word "redesigned" is segmented into "re" and "designed" during preprocessing. The two subwords should share grammartical information among the sentence. Consequently, we will assign them label \textbf{2} and the others label \textbf{0}.

The vector of BPE labels will be input into an embedding layer and a projection layer to obtain BPE embeddings. The BPE embeddings will be added to hidden states before they are transferred into the convolution modules to compute syntactic distances and heights. Concatenating is also considered to combine the two embeddings, but it proves ineffective. We do not introduce BPE embeddings into RoBERTa because of consistency and adaptability from the pre-training stage to fine-tuning.

\subsection{Loss Function}
\label{sec:loss-function}

We also investigate the device of loss function to facilitate grammar induction for the two scenarios studied in our paper.

\citet{loss-mlm-mt} ascertain that the flow of information through Transformer layers is contingent upon the choice of the learning objective. For machine translation (MT), the representation of the input sequence will be refined in the model and transferred from the source language to the target language. In contrast, for masked language modeling (MLM), the information about the context will be rebuilt during the encoding process.

In order to make maximum leverage of grammar induction, we trade off between the two loss functions of MLM and MT with a weighted parameter $\lambda$:

\begin{equation}
    \mathcal{L} = \lambda \cdot \mathcal{L}_{MLM} + (1 - \lambda) \cdot \mathcal{L_{MT}}
\end{equation}
Where the $\lambda$ will be searched for different translation tasks.

Our MLM loss is deployed using the output of the encoder module. We randomly mask part of the input with a special token and restore the masked tokens from the output. The mask rate is set to be 0.15.

In the fine-tuning phase of RoBERTa, we opt not to include the MLM loss. This decision stems from the fact that RoBERTa has already been pre-trained on a vast corpus of data, with MLM serving as its core learning objective. Consequently, the incorporation of MLM loss during fine-tuning does not yield additional benefits. More analyses about the effect of MLM loss are provided in Appendix~\ref{sec:MLM}.

\section{Machine Translation}

We use Transformer (\citealp{transformer}) as our baseline and conduct experiments on six machine translation tasks of three datasets: IWSLT14-De/En, NC11-De/En, and ASPEC-Zh/Ja. Models are trained from scratch and evaluated on each task.

\subsection{Datasets}

% \begin{table*}[t]
% \centering
% \begin{tabular}{lccccccc}
% \toprule
% \multicolumn{1}{l}{\multirow{2}{*}{Models}} &
% \multicolumn{1}{c}{External} &
% \multicolumn{2}{c}{IWSLT14} &
% \multicolumn{2}{c}{NC11} &
% \multicolumn{2}{c}{ASPEC} \\
% \multicolumn{1}{l}{} & Parsers & De$\rightarrow$En & En$\rightarrow$De & De$\rightarrow$En & En$\rightarrow$De & Zh$\rightarrow$Ja & Ja$\rightarrow$Zh \\
% \midrule
% Transformer & w/o. & 34.71 & 28.91 & 27.04 & 25.23 & 47.66 & 34.36 \\
% PASCAL & w. & 34.84 & 29.10 & 27.65 & 25.52 & - & - \\
% ST-NMT* & w. & 35.24 & - & - & - & - & - \\
% Distance Transformer* & w. & 35.74 & 29.28 & 27.67 & \textbf{26.19} & 48.34 & - \\
% \textbf{Ours} & w/o. & \textbf{35.89} & \textbf{29.31} & \textbf{27.69} & 25.83 & \textbf{48.86} & \textbf{34.48} \\
% \bottomrule
% \end{tabular}
% \caption{\label{translation-table}
% BLEU scores on six translation tasks of three datasets. "*" means we report the results of ST-NMT and Distance Transformer from their papers. While "w." means that these methods utilize syntactic information from external parsers, "w/o." means that our method dose does not leverage external parsers.
% }
% \end{table*}

\noindent\textbf{IWSLT14-De/En} The IWSLT14 (International Workshop on Spoken Language Translation) includes the MT track on TED Talks. We use the German (De) and English (En) corpus. We follow the standard pre-processing steps in fairseq (\citealp{fairseq}). The sizes of training, validation and test sets are 160k, 7.3k and 6.8k.

\noindent\textbf{NC11-De/En} The NC11 dataset come from news commentary. We pre-process the dataset following the steps of \citet{nc11}. The sizes of training, validation and test sets are 234k, 2.2k and 3.0k.

\noindent\textbf{ASPEC-Zh/Ja} The ASPEC (Asian Scientific Paper Excerpt Corpus; \citealp{aspec}) dataset is a Chinese (Zh) - Japanese (Ja) scientific paper excerpt corpus. We use the official steps provided by WAT (Workshop on Asian Translation) to pre-process the dataset. The sizes of training, validation and test sets are 672k, 2.1k and 2.1k.

The three datasets we choose are of different scales. Considering the sizes of datasets, IWSLT14 and NC11 tasks could be used to simulate low-resource scenarios. ASPEC, on the other hand, is of large size and provides more source languages.

\subsection{Experiment Settings}

We use six layers of encoder-decoder architecture as the backbone. The number of the convolution layers in our parser is 3, and the dimension of the embedding layer for BPE is set to 256 before input into a projection layer by searching powers of 2 from 2 to 512. We use the syntactic mask in the first encoder layer for all the tasks.

We adopt the inverse square root learning rate scheduler, and the peak learning rate is set to 5e-4, 1e-3 and 3e-4 for IWSLT14, NC11 and ASPEC. The most appropriate values of the weighted parameter $\lambda$ for the loss function are selected by doing a grid search over the range of 0.2 to 0.6 for each task. $\lambda$ is chosen to be 0.47, 0.35, 0.3, 0.45 for IWSLT14, NC11, ASPEC Chinese to Japanese, and Japanese to Chinese respectively. More details of experiment settings are summarized in Appendix~\ref{sec:training-details}.

\subsection{Results}

\begin{table}[t]
\centering
\begin{tabular}{lcc}
\toprule
\multicolumn{1}{l}{\multirow{2}{*}{Models}} &
\multicolumn{2}{c}{IWSLT14} \\
\multicolumn{1}{l}{} & De$\rightarrow$En & En$\rightarrow$De \\
\midrule
Transformer & 34.71 & 28.91 \\
\midrule
\textit{Using external parsers} \\
\quad PASCAL & 34.84 & 29.10 \\
\quad LISA* & 34.97 & 29.06 \\
\quad ST-NMT* & 35.24 & - \\
\quad Distance-Transformer* & 35.74 & 29.28 \\
\midrule
\textit{No external parsers} \\
\quad LPSI*  & 35.05 & - \\
\quad StructFormer  & 32.78 & - \\
\quad \textbf{Ours} & \textbf{35.89} & \textbf{29.31} \\
\bottomrule
\end{tabular}
\caption{\label{iwslt-table}
BLEU scores on IWSLT14-De/En. While PASCAL, ST-NMT and Distance Transformer explicitly leverage external parsers to obtain syntactic information, LPSI, StructFormer and our method introduce self-induced grammar and do not require external parsers. "*" means we report the results of ST-NMT, Distance Transformer and LPSI from their papers. The scores of LISA are the implementation by \citealp{nc11}
}
\end{table}

\begin{table}[t]
\centering
\begin{tabular}{lcc}
\toprule
\multicolumn{1}{l}{\multirow{2}{*}{Models}} &
\multicolumn{2}{c}{NC11} \\
\multicolumn{1}{l}{} & De$\rightarrow$En & En$\rightarrow$De \\
\midrule
Transformer & 27.04 & 25.23 \\
PASCAL & 27.65 & 25.52 \\
LISA* & 27.10 & 25.30 \\
Distance-Transformer* & 27.67 & \textbf{26.19} \\
\textbf{Ours} & \textbf{27.69} & 25.83 \\
\bottomrule
\end{tabular}
\caption{\label{nc11-table}
BLEU scores on NC11-De/En.
}
\end{table}

\begin{table}[t]
\centering
\begin{tabular}{lcc}
\toprule
\multicolumn{1}{l}{\multirow{2}{*}{Models}} &
\multicolumn{2}{c}{ASPEC} \\
\multicolumn{1}{l}{} & Zh$\rightarrow$Ja & Ja$\rightarrow$Zh \\
\midrule
Transformer & 47.66 & 34.36 \\
Distance-Transformer* & 48.34 & - \\
\textbf{Ours} & \textbf{48.86} & \textbf{34.48} \\
\bottomrule
\end{tabular}
\caption{\label{aspec-table}
BLEU scores on ASPEC-Zh/Ja.
}
\end{table}

We take 5 random seeds for each task and average the last 5 checkpoints to be evaluated for each seed. We use the BLEU score to assess the performance of models on the test sets. Our experimental results on the three datasets are presented in Table~\ref{iwslt-table}, Table~\ref{nc11-table} and Table~\ref{aspec-table}. We report the scores of Transformer we implement and other state-of-the-art (SOTA) models, including PASCAL (\citealp{nc11}), LISA \cite{lisa}, ST-NMT (\citealp{st-nmt}) and Distance Transformer (\citealp{distance-tranformer}), which are enhanced by external parsers, as well as LPSI \cite{LPSI}, which unsupervisedly induces latent phrase structures and incorporate them into the attention mechanism. We also replace the Transformer encoder with StructFormer \cite{structformer}, which is an encoder-based model for grammar parsing, and train the model on IWSLT14 De$\rightarrow$En.

Tables demonstrate that our method obtains SOTA results in most translation tasks. It consistently outperforms the vanilla Transformer across all six tasks, with particularly notable improvements observed in IWSLT14 German to English and ASPEC Chinese to Japanese tasks. It surpasses other external-parser-enhanced methods and the unsupervised method LPSI and StructFormer on IWSLT14-De/En.

It is worth noting that our method achieves scores comparable to the reported outcomes of Distance Transformer (\citealp{distance-tranformer}). They rely on the external parser to generate constituency grammar information so as to enhance the self-attention mechanism in Transformer. Moreover, when we reproduce their results, we find that their model benefits from attention dropout, and they use the best checkpoint at inference time. Compared with their supervised Distance Transformer, our unsupervised method learns to comprehend grammar structures without any additional knowledge or external tools.

In addition, it can be observed that all the other external-parser-enhanced methods are mainly applied in German and English since the parsers they leveraged are primarily developed for these two languages. Their models are affected by external parsers and limited in the diversity of languages. In contrast, our method internalizes grammar induction within Transformer for downstream tasks without the requirement for additional syntactic annotations. As a result, our approach can be applied across a wider range of languages.

\begin{table*}
\centering
\resizebox{2.05\columnwidth}{!}{
\begin{tabular}{lcccccc}
\toprule
\multicolumn{1}{l}{\multirow{2}{*}{Models}} &
\multicolumn{2}{c}{IWSLT14} &
\multicolumn{2}{c}{NC11} &
\multicolumn{2}{c}{ASPEC} \\
\multicolumn{1}{l}{} & De$\rightarrow$En & En$\rightarrow$De & De$\rightarrow$En & En$\rightarrow$De & Zh$\rightarrow$Ja & Ja$\rightarrow$Zh \\
\midrule
Transformer & 34.71$\pm$0.12 & 28.91$\pm$0.14 & 27.04$\pm$0.10 & 25.23$\pm$0.17 & 47.66$\pm$0.25 & 34.36$\pm$0.09 \\
Ours (-BPE) & 35.83$\pm$0.07 & 29.19$\pm$0.12 & 27.39$\pm$0.11 & 25.27$\pm$0.63 & 48.78$\pm$0.23 & \textbf{34.70$\pm$0.12} \\
\textbf{Ours (+BPE)} & \textbf{35.89$\pm$0.18} & \textbf{29.31$\pm$0.13} & \textbf{27.69$\pm$0.17} & \textbf{25.83$\pm$0.26} & \textbf{48.86$\pm$0.14} & 34.48$\pm$0.27 \\
\bottomrule
\end{tabular}}
\caption{\label{bpe-table}
Effect of BPE embeddings. "-BPE" means using our method without BPE embeddings, while "+BPE" means using our method with BPE embeddings.
}
\end{table*}                             

\subsection{Effect of BPE Embddings}

To assess the effect of BPE embeddings we have designed, we compare the performance of our method with and without the incorporation of BPE embeddings.

\begin{table}[]
\centering
\begin{tabular}{lcccc}
\toprule
\multicolumn{1}{l}{\multirow{2}{*}{tasks}} &
\multicolumn{2}{c}{De$\rightarrow$En} & \multicolumn{2}{c}{En$\rightarrow$De} \\
\multicolumn{1}{l}{} & -BPE & \textbf{+BPE} & -BPE & \textbf{+BPE} \\
\midrule
precision & 35.55 & \textbf{35.70} & 25.63 & \textbf{31.96}  \\
recall & 34.13 & \textbf{34.27} & 24.59 & \textbf{30.66}  \\
F1 & 34.83 & \textbf{34.97} & 25.10 & \textbf{31.30}  \\
\bottomrule
\end{tabular}
\caption{Parsing performance on IWSLT14.}
\label{parsing-table}
\end{table}

The average scores and standard deviations presented in Table \ref{bpe-table} indicate that while our BPE embeddings have a modest influence on machine translation, they do offer a positive contribution across all tasks except the ASPEC Japanese to Chinese task. It is observable that BPE embeddings are particularly beneficial in the context of English to German translations. This enhanced utility is likely attributed to the fact that the grammar status sharing among congenetic subwords is more compatible with English grammar. Regardless of the presence of BPE embeddings, our method shows consistent advancements over Transformer across machine translation tasks, showcasing the robustness and adaptability of our approach.

\begin{figure}[!h]
    \centering
        \subfigure[The ground truth tree obtained by Stanford CoreNLP.]{   
        % \hspace{-0.1cm}
        \includegraphics[width=1.0\linewidth]{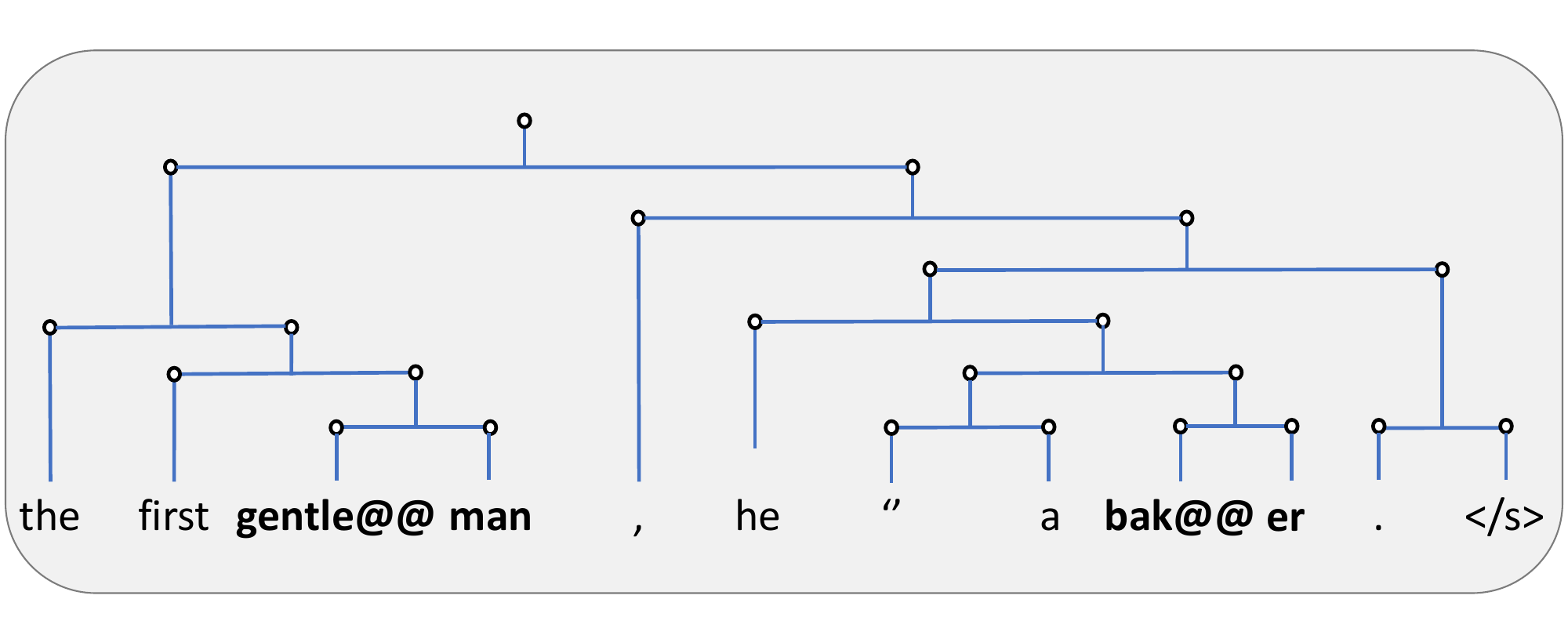}  
        \label{gt-tree}
        }
        \subfigure[The constituency tree obtained by using our method without BPE embeddings.]{
        % \hspace{-0.1cm}
        \includegraphics[width=1.0\linewidth]{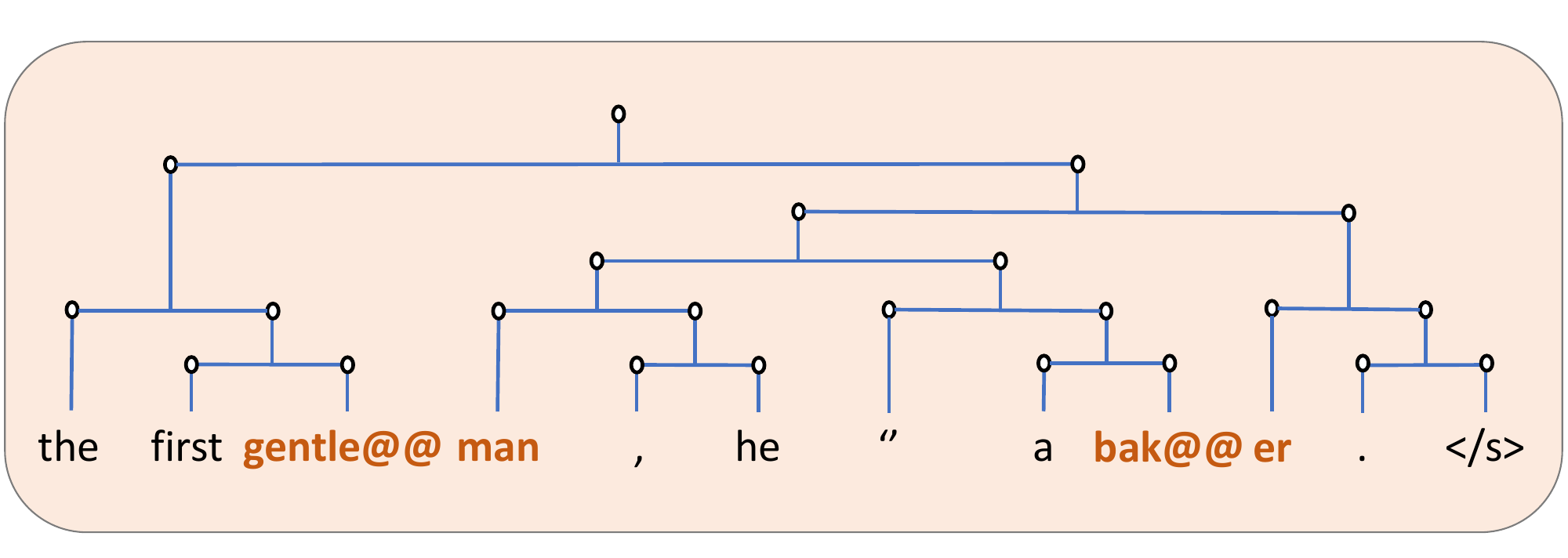}
        \label{nobpe-tree}
        }
        \subfigure[The constituency tree obtained by using our method with BPE embeddings.]{   
        % \hspace{-0.1cm}
        \includegraphics[width=1.0\linewidth]{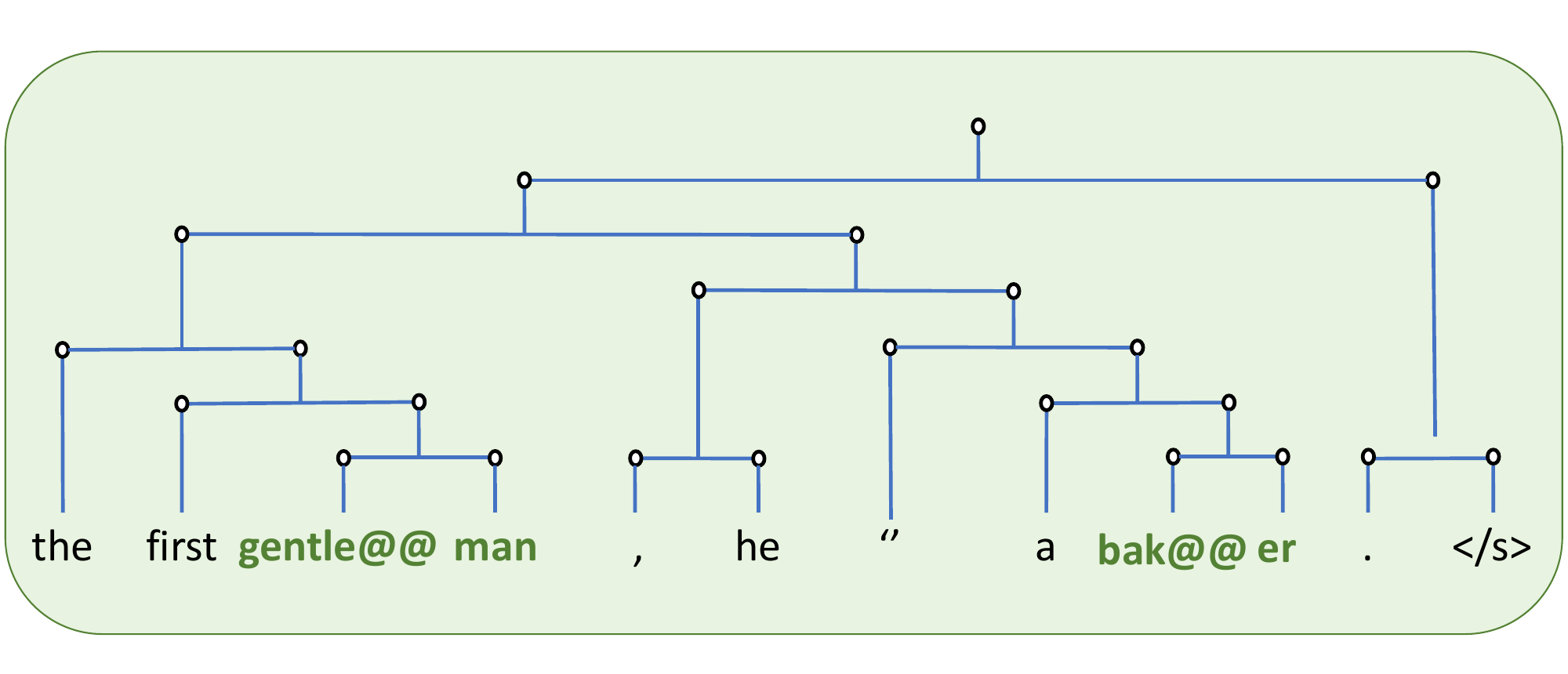}  
        \label{bpe-tree}
        }
    % \vspace{-5cm}
    \caption{The constituency tree for the example sentence by using (a) Stanford CoreNLP, (b) our method without BPE embeddings, and (c) our method with BPE embeddings. "@@" in "gentle@@" and  "bak@@" is the sign of BPE segmentation. }
    \label{tree-fig}
\end{figure}

\subsection{Parsing Performance}
\label{sec:parsing}

Furthermore, we implement the distance-to-tree algorithm (\citealp{distance2tree}) to reconstruct the hierarchical structure of a constituency tree, leveraging the syntactic distances generated during the parsing process. Following the definition of syntactic distance, the two words with the smallest distance will be merged into a tree first. The sequence of tokens, along with their syntactic distances, will be converted into a binary tree. The transformation algorithm can be referred to in Appendix~\ref{sec:transformation-algorithm}.

To assess the parsing performance of our method, we utilize the validation set of IWSLT14-De/En. The parser of Stanford CoreNLP\footnote{https://nlp.stanford.edu/software/segmenter.shtml} is employed to generate constituency trees as reference.\footnote{Stanford CoreNLP was also leveraged by previous work \cite{constituency-nmt} to generate standard parses for samples of machine translation tasks. These generated parses served as references for evaluating parsing performance during translation.} Precision, recall, and F1 scores are computed to evaluate the trees generated by our method, with the results presented in Table \ref{parsing-table}.

% \begin{table*}[ht]
% \centering
% \resizebox{2.05\columnwidth}{!}{
% \begin{tabular}{lcccccccccc}
% \toprule
% \multirow{2}{*}{Models} & External & CoLA & SST-2 & MNLI & QNLI & RTE & MRPC & QQP & STS-B & \multirow{2}{*}{Avg} \\
%  & Parsers & (mc) & (acc) & m/mm(acc) & (acc) & (acc) & (F1) & (F1) & (pc) &  \\
% \midrule
% BERT* & w/o. & 52.1 & 93.5 & 84.6/83.4 & 90.5 & 66.4 & 88.9 & 71.2 & 85.8 & 79.6 \\

% POS-BERT* & w. & 52.9 & 93.9 & 84.4/83.3 & 90.4 & 66.9 & 88.8 & 71.4 & 85.5 & 79.7 \\

% RoBERTa & w/o. & 61.3 & 95.2 & 87.6/87.0 & 92.8 & 75.8 & 92.9 & 88.9 & 90.9 & 85.8 \\

% SLA* & w. & 60.0 & 93.3 & -/- & 91.4 & 67.8 & - & - & 89.9 & - \\

% Syntax-RoBERTa* & w. & 63.3 & \textbf{96.1} & 87.8/85.7 & \textbf{94.3} & \textbf{81.2} & 88.5 & 88.5 & 89.9 & 86.1 \\

% SynCLM* & w. & \textbf{65.3} & 95.1 & 87.2/- & 93.0 & 80.1 & \textbf{93.7} & 88.9 & 90.8 & - \\

% \textbf{Ours} & w/o. & 62.2$^{\uparrow}$ & 95.0 & \textbf{87.7$^{\uparrow}$/87.5$^{\uparrow}$} & 93.3$^{\uparrow}$ & 79.8$^{\uparrow}$ & 92.8 & \textbf{89.1$^{\uparrow}$} & \textbf{90.9} & \textbf{86.5$^{\uparrow}$} \\

% \bottomrule
% \end{tabular}}
% \caption{\label{glue-table}
% Scores on GLUE benchmark. The evaluation metrics are listed below the task names, where "mc" denotes Matthews correlation coefficient, "acc" denotes accuracy, "F1" denotes F1 score, and "pc" denotes Pearson correlation coefficients. The MNLI dataset has 2 versions of test sets represented as "m" and "mm". The scores with a superscript "$\uparrow$" denote our method outperforms the vanilla RoBERTa. 
% }
% \end{table*}

\begin{table*}[ht]
\centering
\resizebox{2.05\columnwidth}{!}{
\begin{tabular}{lccccccccc}
\toprule
\multirow{2}{*}{Models} & CoLA & SST-2 & MNLI & QNLI & RTE & MRPC & QQP & STS-B & \multirow{2}{*}{Avg} \\
 & (mc) & (acc) & m/mm(acc) & (acc) & (acc) & (F1) & (F1) & (pc) &  \\
\midrule
RoBERTa & 61.3 & 95.2 & 87.6/87.0 & 92.8 & 75.8 & 92.9 & 88.9 & \textbf{90.9} & 85.8 \\
\midrule
\textit{Using external parsers} \\
\quad SLA* & 60.0 & 93.3 & -/- & 91.4 & 67.8 & - & - & 89.9 & - \\

\quad Syntax-RoBERTa* & 63.3 & \textbf{96.1} & 87.8/85.7 & \textbf{94.3} & \textbf{81.2} & 88.5 & 88.5 & 89.9 & 86.1 \\

\quad SynCLM* & \textbf{65.3} & 95.1 & 87.2/- & 93.0 & 80.1 & \textbf{93.7} & 88.9 & 90.8 & - \\

\midrule
\textit{No external parsers} \\
\quad \textbf{Ours} & 62.2$^{\uparrow}$ & 95.0 & \textbf{87.7$^{\uparrow}$/87.5$^{\uparrow}$} & 93.3$^{\uparrow}$ & 79.8$^{\uparrow}$ & 92.8 & \textbf{89.1$^{\uparrow}$} & \textbf{90.9} & \textbf{86.5$^{\uparrow}$} \\

\bottomrule
\end{tabular}}
\caption{\label{glue-table}
Scores on GLUE benchmark. The evaluation metrics are listed below the task names, where "mc" denotes Matthews correlation coefficient, "acc" denotes accuracy, "F1" denotes F1 score, and "pc" denotes Pearson correlation coefficients. The MNLI dataset has 2 versions of test sets represented as "m" and "mm". The scores with a superscript "$\uparrow$" denote our method outperforms the vanilla RoBERTa. 
}
\end{table*}

The F1 scores of constituency trees generated by using our method achieve 34.97 and 31.30 for the source languages of German and English, respectively. While the parsing performance may not match that of an expert-supervised grammar parser, with such moderate parsing accuracy, our method could achieve reasonable improvement in machine translation. 

Additionally, it can be observed from Table \ref{parsing-table} that the scores using BPE embeddings surpass those without BPE embeddings. In Figure \ref{tree-fig}, we illustrate constituency trees generated by the three different methods for an example sentence. Compared to the ground truth, the hierarchical structure of the constituency tree generated by using our method appears quite plausible. It suggests that BPE embeddings enhance the model's ability to capture BPE segmentations, which proves that the grammar status sharing among congenetic subwords is practical and valid.

\section{Language Understanding}

We deploy our grammar induction method on RoBERTa (\citealp{roberta}) at the stage of fine-tuning, and evaluate its performance in the GLUE benchmark \cite{glue}.

\subsection{Datasets}
The GLUE (General Language Understanding Evaluation) benchmark is a collection of datasets for evaluating the natural language understanding performance of models. It consists of (1) single-sentence classification tasks: CoLA (Corpus of Linguistic Acceptability) and SST-2 (Stanford Sentiment Treebank); (2) similarity and paraphrase tasks: MRPC (Microsoft Research Paraphrase Corpus), QQP (Quora Question Pairs) and STS-B (Semantic Textual Similarity Benchmark); (3) inference tasks: MNLI (Multi-Genre Natural Language Inference Corpus), QNLI (Stanford Question Answering Dataset) and RTE (Recognizing Textual Entailment).

\subsection{Experiment Settings}

RoBERTa is a well and robustly pre-trained model with a strong ability for language understanding. It is essentially a Transformer model with only encoder layers. We use RoBERTa-base as our backbone. It has 12 encoder layers and 12 attention heads in each layer. We load official checkpoints of pre-trained RoBERTa. Details of the experimental settings are set following fairseq (\citealp{fairseq}).

For our grammar-aware RoBERTa, the number of convolution layers in our parser is 2, and we use the syntactic mask in the first three encoder layers. We fine-tune and evaluate the model on each task of GLUE.

\subsection{Results}

The results on the GLUE benchmark are presented in Table \ref{glue-table}. The best checkpoints are saved during training, and we use standard evaluation metrics for each task. We reproduce the vanilla RoBERTa and our grammar-enhanced RoBERTa in the same environment and report the scores in the table.

As shown in the table, our method outperforms the vanilla RoBERTa in the majority of GLUE tasks and achieves the highest average score. Remarkably, without relying on any additional syntactic annotations, our method has achieved improvements that are more consistent when compared to other syntax-enhanced models, including SLA (\citealp{sla-bert}), Syntax-RoBERTa (\citealp{syntax-bert}) and SynCLM (\citealp{synclm}). To introduce syntactic information into pre-trained models, all of these methods resort to external parsing models.

When integrated with our method for self-inducing grammatical knowledge, RoBERTa obtains improvements across six of the nine test sets in GLUE, with the scores on the remaining three test sets being nearly identical to those of the baseline RoBERTa. More specifically, it can be inferred from the table that pre-trained language models have exhibited such an impressive capability of language understanding that the integration of external parsers may not bring uniform enhancements across all tasks. They even hurt the performance in certain tasks, as evidenced by the performance of SLA on RTE, Syntax-RoBERTa on MRPC, and SynCLM on MNLI. Nonetheless, our approach performs comparably well in single-sentence classification tasks, and similarity and paraphrase tasks. It particularly excels in inference tasks, with a notable improvement observed in the RTE task.

\section{Related Work} 
There are generally two ways for syntactic knowledge-based language comprehension.

\paragraph{Explict syntax enhancement}
One way is to incorporate syntactic information explicitly by using a high-quality external parser. PASCAL \cite{nc11} builds up a syntax-guided localized attention mask where each token's attention range is a Gaussian distribution centered by its dependency head.  Distance Transformer \cite{distance-tranformer} utilizes the relative magnitude of syntactic distance from the constituency tree to build up a syntactic local range. ST-NMT \cite{st-nmt} slices the constituency tree at some depth and gets a list of constituents as labels. A new Transformer encoder is learned to predict this structure, and its output is integrated with the original encoder output. SEPREM \cite{seprem} also focuses on pretraining architecture and defines a prior attention weight distribution by normalizing the inverse of the token's distance in the dependency structure. However, all of these methods require sophisticated external parsers to generate dependency or constituency syntax trees.

%  ON-LSTM\cite{shen2018ordered} designs a way of ordering the neurons in LSTM architecture to latently induce a hierarchy of constituents.

\paragraph{Implicit syntax enhancement}
The other way is to implicitly induce the latent grammar structure by learning from data distribution. PRPN \cite{PRPN} enhances RNN neural language modeling by simultaneously learning a CNN-based parsing network, which could induce latent constituency structure represented by syntactic distance. Recent works have focused more on attention mechanisms. StructFormer \cite{structformer} introduces a parser layer that can induce dependency and constituency syntax simultaneously and integrates induced dependency structure into self-attention in a differentiable way. LPSI \cite{LPSI} induces latent phrase structure in an unsupervised fashion and integrates them into the multi-head attention mechanism. SyncAttn \cite{syncattn} enables neural OpenIE to induce the latent syntactic structure and adopt multi-view learning to capture multiple relationships from constituency and dependency graphs. Nevertheless, very little work pays attention to leveraging implicit induction in downstream tasks.

\paragraph{Syntax enhancement of pre-trained models}
In recent years, people also studied to introduce grammar induction into pre-trained models like BERT \cite{bert} and Roberta \cite{roberta} instead of training from scratch. POS-BERT \cite{pos-bert} leverages POS tag information apart from the syntactic structure. SLA \cite{sla-bert} prohibits two tokens from attending to each other once their distance in the dependency tree exceeds some boundary, and it improves the fine-tuning performance on several GLUE tasks. Syntax-Roberta \cite{syntax-bert} develops a framework that can be easily plugged into an arbitrary pre-trained checkpoint, which automatically underlines the most relevant syntactic knowledge for each downstream task. SynCLM \cite{synclm} proposes a syntax-guided contrastive learning method where phrase-guided and tree-guided contrastive objectives based on constituency and dependency structures are optimized in the pre-training stage to help the language model capture rich syntactic knowledge in its representations. However, these methods resort to external parsers to obtain syntactic information, which could not work well with pre-trained models across downstream tasks.

\section{Conclusion}

In this paper, we study the utilization of grammar induction for language understanding and generation. We force the model to be trained with the self-induced grammar without external parsers or annotations, which is a more general approach. Moreover, our method is efficient in both the from-scratch and pre-trained scenarios. Through assessments across various machine translation and language comprehension tasks, we validate the efficacy of our grammar induction technique, showcasing substantial and consistent enhancements. These results underscore the tangible benefits that can be achieved through the incorporation of grammar induction techniques. Our research offers promising insights into the practical application of induced grammar in downstream tasks.

\section*{Limitations}
The performance of the grammar-aware method may be constrained in large-scale datasets and well-trained models. As shown in our paper, none of the grammar-aware methods can achieve consistent improvements on all the tasks of GLUE. This phenomenon could be explained by the adequacy of model training. Since RoBERTa has been pre-trained over a tremendous amount of data, it has already done well in language understanding. But it occurs to us whether pre-trained models or large language models have some form of syntax parsing modules inside them.

The parsing performance of our parser is a little limited. We found that the quality of constituency trees generated from the syntactic distance in our parser is far from that of well-designed and trained supervised constituency parsers. However, our method does make neural network models more interpretable. More appropriate optimization methods could be explored to make grammar induction more
effective.

\section*{Ethics Statements}

Our work pertains to neural machine translation and fine-tuning of pre-trained language models to introduce grammar induction into Transformer. In this work, we use only publicly available data and artifacts.

\section*{Acknowledgement}

This work was sponsored by the National Natural Science Foundation of China (NSFC) grant (No.62106143), and National Key Research and Development Program of China (No. 2023ZD0121402) .

\bibliography{custom}

\appendix

\begin{table*}[t]
\centering
\begin{tabular}{lcccc}
\toprule
\multicolumn{1}{l}{\multirow{2}{*}{Models}} &
\multicolumn{2}{c}{IWSLT14} &
\multicolumn{2}{c}{NC11} \\
\multicolumn{1}{l}{} & De$\rightarrow$En & En$\rightarrow$De & De$\rightarrow$En & En$\rightarrow$De \\
\midrule
Transformer & 34.71 & 28.91 & 27.04 & 25.23 \\
Ours (-MLM) & 34.41 & 28.44 & 26.24 & 25.26 \\
\textbf{Ours (+MLM)} & \textbf{35.89} & \textbf{29.31} & \textbf{27.69} & \textbf{25.83} \\
\bottomrule
\end{tabular}
\caption{\label{mlm-mt}
Effect of MLM loss on machine translation. "-MLM" means using our method without MLM loss, while "+MLM" means using our method with MLM loss.
}
\end{table*} 

\begin{table*}[ht]
\centering{
\begin{tabular}{lccccccccc}
\toprule
\multirow{2}{*}{Models} & CoLA & SST-2 & MNLI & QNLI & RTE & MRPC & QQP & STS-B & \multirow{2}{*}{Avg} \\
 & (mc) & (acc) & m/mm(acc) & (acc) & (acc) & (F1) & (F1) & (pc) &  \\
\midrule
RoBERTa & 61.3 & \textbf{95.2} & 87.6/87.0 & 92.8 & 75.8 & \textbf{92.9} & 88.9 & \textbf{90.9} & 85.8 \\

\textbf{Ours (-MLM)} & \textbf{62.2} & 95.0 & 87.7/87.5 & \textbf{93.3} & \textbf{79.8} & 92.8 & \textbf{89.1} & \textbf{90.9} & \textbf{86.5} \\

Ours (+MLM) & 62.1 & 95.0 & \textbf{88.0/87.5} &	93.1 & 75.8 & 92.7 & 88.7 & 89.4 & 85.8 \\

\bottomrule
\end{tabular}}
\caption{\label{mlm-glue}
Effect of MLM loss on GLUE.
}
\end{table*}

\section{Estimation of Dependency Distribution}
\label{sec:structformer}

The estimation function $\mathcal{F_P}(\cdot)$ in Equation~\ref{eq:estimation} leverages the two-stage method by \citet{structformer} to calculate the estimation of the dependency distribution among tokens. We utilize it to generate the syntactic mask with syntactic distance and height. We describe the calculation procedure in detail.

To estimate the probability $p_{\mathbf{D}}(w_j|w_i)$ that the $j$-th token $w_j$ is the parent of the $i$-th token $w_i$, they decompose $p_{\mathbf{D}}(w_j|w_i)$ into two factors: 
\begin{equation}
p_{\mathbf{D}}(j \mid i)=
\sum_{[l, r]} p_{\mathbf{P r}}(j \mid[l, r]) p_{\mathbf{C}}([l, r] \mid i)
\end{equation}
where $p_{\mathbf{P r}}(j \mid[l, r])$ denotes the probability that $w_j$ is the root of $w_{l\dots r}$. It can be parameterized by syntactic height:
\begin{equation}
p_{\mathbf{P r}}(j \mid[l, r])=
\frac{\exp \left(h_j\right)}{\sum_{l \leq k \leq r} \exp \left(h_k\right)}
\end{equation}
And $p_{\mathbf{C}}([l, r] \mid i)$ denotes the probability that $w_{l\dots r}$ is the smallest legal constituent $C(w_i)$ for $w_i$:
\begin{equation}
p_{\mathbf{C}}([l, r] \mid i)=
p_{\mathbf{L}}(l \mid i) p_{\mathbf{R}}(r \mid i)
\end{equation}
where $p_{\mathbf{L}}(l \mid i)$ denotes the probability that $l$ is the left margin of $C(w_i)$ and $p_{\mathbf{R}}(r \mid i)$ stands for the right margin.

The distribution is parameterized  that the $l$-th token $w_l$ is inside $C(w_i)$ with the probability $p(l \in C(w_i))$ that $h_i$ is larger than the maximum distance $\tau$ between $l$ and $i$:
\begin{equation}
p\left(l \in \mathbf{C}\left(w_i\right)\right)
=\sigma\left(\left(h_i-\max \left(\tau_l, \ldots, \tau_{i-1}\right)\right) / \mu\right)
\end{equation}
where $\sigma$ is the sigmoid function and $\mu$ is a learnable temperature term. 

The event $l \in C(w_i)$ consists of 2 cases: $l \in C(w_i) \cap l-1 \in C(w_i)$ and $l \in C(w_i) \cap l-1 \notin C(w_i)$.

For the first case, if $l-1 \in C(w_i)$, we must have $l \in C(w_i)$. So the probability can be measured with $p(l-1 \in C(w_i))$. For the second case, if $l \in C(w_i)$ and $l-1 \notin C(w_i)$, it means that $l$ is the left margin of $C(w_i)$.

Therefore, the probability $p_{\mathbf{L}}(l \mid i)$ that $l$ is the left margin of $C(w_i)$ can be derived by:
\begin{equation}
p_{\mathbf{L}}(l \mid i)=p\left(l \in \mathbf{C}\left(w_i\right)\right)-p\left(l-1 \in \mathbf{C}\left(w_i\right)\right)
\end{equation}

And $p_{\mathbf{R}}(r \mid i)$ can be derived by:
\begin{equation}
p_{\mathbf{R}}(r \mid i)=p\left(r-1 \in \mathbf{C}\left(w_i\right)\right)-p\left(r \in \mathbf{C}\left(w_i\right)\right)
\end{equation}

The procedure goes through the sentence to estimate the syntactic dependency of each token on the others.

\section{Experiment Details}
\label{sec:training-details}

We perform all the training on 2 RTX3090 GPUs for IWSLT14 and NC11, 2 80G-A100 for ASPEC, and 1 40G-A100 for GLUE. For our method with BPE embeddings, each training took about 2.5 hours for IWSLT14, 3 hours for NC11, and 50 hours for ASPEC. For our method without BPE embeddings, each training took about 2 hours for IWSLT14, 2.5 hours for NC11, and 29 hours for ASPEC. We use Adam optimizer with $\beta_{1}=0.9, \beta_{2}=0.98$. Cross entropy is the default loss function with label smoothing of 0.1 and weight decay of 0.0001. The batch size ($GPU\_num \times Update\_num \times Max\_tokens$) is $2 \times 1 \times 4096$, $2 \times 2 \times 8192 $ and $ 2 \times 1 \times 8192$ respectively.

The numbers of attention heads in each layer and dimensions of the fully connected layer are 4 and 1024 for IWSLT14 and NC11, 8 and 2048 for ASPEC.

We employ beam search at inference time. Following the settings of previous work \citealp{distance-tranformer}, the beam size and length penalty are set to 5 and 1.0 for IWSLT14 and ASPEC, and set to 4 and 0.6 for NC11.

The number of model parameters is about 49M for IWSLT14, 57M for NC11, and 108M for ASPEC, with an increase of 9M. For grammar-aware RoBERTa on GLUE, the fine-tuning took about 7 hours and the number of model parameters is about 135M with an increase of 10M.

We also see increased numbers of parameters in other syntax-enhanced models like SLA \cite{sla-bert}, ST-NMT \cite{st-nmt}, LPSI \cite{LPSI} and Syntax-Roberta \cite{syntax-bert}, while they do not report statistics in their papers.

\section{Effect of MLM loss}
\label{sec:MLM}

We conduct experiments to study the effect of MLM loss on our method. The results on machine translation tasks and GLUE are shown in Table~\ref{mlm-mt} and Table~\ref{mlm-glue}.

Experiment results demonstrate that MLM loss has a different effect on models' grammar integration in the two scenarios. As discussed in Section~\ref{sec:loss-function}, the language modeling objective contributes to the hierarchical generalization of Transformers. Consequently, our grammar induction works well with MLM loss to train Transformers from scratch on machine translation tasks. Nevertheless, it does not benefit from MLM loss when fine-tuning a pre-trained model RoBERTa. One possible reason is that learning to induce grammar with MLM loss might influence hierarchical generalization inside the model, which has been developed by MLM loss in the pre-training of RoBERTa. It deserves further work to study the relation between language modeling and grammar induction during different stages of training.

\section{Transformation Algorithm}
\label{sec:transformation-algorithm}

The algorithm we implemented in Section~\ref{sec:parsing} is to transform the syntactic distances of a sentence into a constituency tree.

The transformation algorithm is designed with a top-down merging method. It divides the input sequence $w$ into two subsequences at the split point $i$ whose syntactic distance $\tau_i$ is the largest in the sequence. Then the subtrees of the two subsequences are merged into a whole.

\begin{algorithm}[!t]
\caption{Syntactic Distance to Constituency Tree} 
\label{algorithm} 
    \hspace*{0.02in} {\bf Input:}
    Syntactic distance $\tau_{1\dots n-1}$, sentence $w_{1\dots n}$ \\
    \hspace*{0.02in} {\bf Output:} 
    Constituency Tree $T$ \\
    \hspace*{0.02in} {\bf Function}
    Tree($\tau, w$)
    \begin{algorithmic}[1]
    \IF{$\tau = []$} 
    \STATE $T \leftarrow$ Leaf($w$)
    \ELSE 
    \STATE $i \leftarrow$ argmax$_i$($\tau$)
    \STATE $node_l \leftarrow$ Tree($\tau_{1\dots i-1}, w_{1\dots i}$)
    \STATE $node_r \leftarrow$ Tree($\tau_{i+1\dots n-1}, w_{i+1\dots n}$)
    \STATE $T \leftarrow$ Node($node_l$, $node_r$)
    \ENDIF 
    \RETURN $T$
    \end{algorithmic}
\end{algorithm}

% \begin{table*}
% \centering
% \begin{tabular}{lcccccc}
% \hline
% \multicolumn{1}{l}{\multirow{2}{*}{Models}} &
% \multicolumn{2}{c}{IWSLT14} &
% \multicolumn{2}{c}{NC11} &
% \multicolumn{2}{c}{ASPEC} \\
% \multicolumn{1}{l}{} & De$\rightarrow$En & En$\rightarrow$De & De$\rightarrow$En & En$\rightarrow$De & Zh$\rightarrow$Ja & Ja$\rightarrow$Zh \\
% \hline
% Transformer & 34.71$\pm$0.12 & 28.91$\pm$0.14 & 27.04$\pm$0.10 & 25.23$\pm$0.17 & 47.66$\pm$0.25 & 34.36$\pm$0.09 \\
% -BPE & 35.83$\pm$0.07 & 29.19$\pm$0.12 & 27.39$\pm$0.11 & 25.27$\pm$0.63 & 48.78$\pm$0.23 & \textbf{34.70$\pm$0.12} \\
% \textbf{+BPE} & \textbf{35.89$\pm$0.18} & \textbf{29.31$\pm$0.13} & \textbf{27.69$\pm$0.17} & \textbf{25.83$\pm$0.26} & \textbf{48.86$\pm$0.14} & 34.48$\pm$0.27 \\
% \hline
% \end{tabular}
% \caption{\label{bpe-table}
% Effect of BPE embeddings. "-BPE" means using our method without BPE embeddings, while "+BPE" means using our method with BPE embeddings.
% }
% \end{table*}      

\end{document}